\documentclass[11pt]{article}

\usepackage[preprint]{acl}

\usepackage{times}
\usepackage{latexsym}
\usepackage{booktabs}
\usepackage{tabularx}
\usepackage{multirow}
\usepackage[most]{tcolorbox}
\usepackage[dvipsnames]{xcolor}

\newtcolorbox{promptbox}[1][]{
    breakable,
    colback=gray!5,
    colframe=gray!50,
    boxrule=0.5pt,
    arc=2pt,
    left=5pt,
    right=5pt,
    top=5pt,
    bottom=5pt,
    fonttitle=\bfseries,
    title=#1,
}
\usepackage[T1]{fontenc}

\usepackage[utf8]{inputenc}

\usepackage{microtype}

\usepackage{inconsolata}

\usepackage{graphicx}
\usepackage[export]{adjustbox}
\usepackage{enumitem}
\usepackage{linguex}

\newcommand{\propdirect}{\textsc{DPSal}}
\newcommand{\propaligned}{\textsc{SaPSal}}
\newcommand{\propgraded}{\textsc{GaPSal}}

\title{The Sum of All Summaries: \\
\textsc{PropAlign} for Graded Proposition Salience Evaluation}

\author{Lin Ai$^{1}$ \quad Emma Thronson$^{1}$ \quad Katherine Conhaim$^{1}$ \quad Amir Zeldes$^1$ \\
$^1$Department of Linguistics, Georgetown University, USA \\
{\small 
    \textbf{Correspondence:} \href{mailto:la987@georgetown.edu}{\texttt{la987@georgetown.edu}}}
}

\begin{document}
\maketitle
\begin{abstract}
Identification and ranking of salient propositions in documents is indirectly involved in many AI tasks, but few papers define proposition salience scores clearly, make diverse data available, or evaluate models directly on the task. This paper proposes \textsc{PropAlign}, a multiple-summary-based approach for propositional salience score estimation, and releases \textsc{PropAlignBank}, an accompanying dataset for the method with 138K alignments of over 35K propositions. We compare several approaches to salient proposition extraction in English across a broad range of spoken and written genres, and demonstrate that naively asking models for judgments leads to inconsistencies. By contrast, graded, summary-based metrics are more stable. We further observe that while single models perform worse than humans, an ensemble of models incorporating linguistic features approaches manual annotation quality ceilings. In addition, analyses of the model outputs suggest insights into how models can extract salient information with and without the use of summaries.\footnote{\textsc{PropAlignBank} and our code will be made publicly available upon acceptance.}
\end{abstract}

\section{Introduction}

\begin{figure}[t!hb]
  \centering
  \includegraphics[width=\columnwidth, frame]{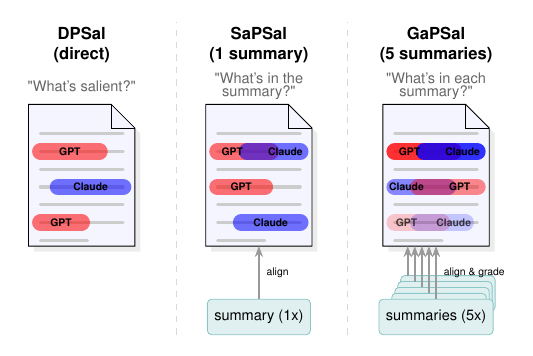}
\caption{Asking models for salient text is more subjective and less consistent than asking what appears in a summary. Multiple summaries can compensate for summary subjectivity and yield gradient salience scores, and model judgments overlap more on the right.}
  \label{fig:topright}
  \vspace{-10pt}
\end{figure}

Identifying the most salient information in a document and ranking information for salience are important tasks in information retrieval, automatic summarization, and computational social science \cite{wang-etal-2020-heterogeneous,bar-haim-etal-2020-quantitative,le-luu-2024-extractive,dong-etal-2025-termdiffusum}. A major challenge in approaching salience in natural language is that definitions are very general and appeal to subjective judgments, which make it hard for both humans and models to agree on `the' correct answer, or to place salience judgments on a reliable ratio scale.

Motivated by the need for a stable operationalization of proposition salience, in this paper we propose \textsc{PropAlign}, an approach that utilizes summarization-based salience metrics previously used for Salient Entity Extraction (SEE; \citealt{bullough-etal-2024-predicting,lin-zeldes-2024-gumsley}). The framework relies on the idea that if something is salient in a document, it will be difficult to summarize the document without mentioning it \cite{dunietz-gillick-2014-new}. However, because this approach is susceptible to the inherent subjectivity and variability of individual summaries and yields only a binary judgment of salience, we adopt a more nuanced approach that uses multiple summaries to mitigate individual idiosyncrasies and yield a graded salience metric. We refer to directly querying for salience as \textsc{DPSal} (direct proposition salience), querying based on appearance in a single summary as SaPSal (Summary-aligned Proposition Salience), and using multiple summaries as GaPSal (Graded-aligned Proposition Salience, see Figure \ref{fig:topright} and Section \ref{sec:settings}).

A further question that frameworks seeking to operationalize proposition salience must contend with is the unit of analysis. While documents are often divided into sentences to be ranked for importance, sentences can be quite long, and it is possible for only parts of a sentence to be relevant \cite{zhou-etal-2020-level,PengGeWang2023}. \citet{zeldes-etal-2026-worth} found that among the 20\% of sentences that contained salient information, half of them included non-salient clauses, motivating a focus on smaller units. Though various possibilities exist, \citet{wu-etal-2023-edu} found that Elementary Discourse Units (EDUs) as defined in Rhetorical Structure Theory \cite{MannThompson1988}, which correspond roughly to clauses or predications, are an optimal unit for fine-grained content extraction. Because EDUs have been studied extensively and have clear segmentation guidelines \cite{CarlsonMarcu2001} and abundant publicly available data, for the remainder of this paper, we will use EDUs as the operationalization for propositions.

The main contributions of this paper are:

\begin{itemize}[itemsep=0.1em]
    \item We present \textsc{PropAlign}, a method and task for salient proposition extraction
    \item We release \textsc{PropAlignBank}, the first large scale dataset for graded summary-based proposition salience scoring
    \item Our annotations include tens of thousands of manual alignments for evaluation
    \item We evaluate three open-weight and frontier LLMs on several versions of the task and create an ensemble to produce additional silver standard data
\end{itemize}

\section{Related work}

\subsection{Proposition salience extraction}

Little previous work has targeted salience extraction outside of downstream tasks such as extractive summarization (see Section \ref{sec:extractive}). The closest work in terms of goals is \citet{liu-etal-2018-automatic}, who attempted to extract salient events from the NYT corpus \cite{Sandhaus2008}. However, their approach considered only a restricted subset of verbal predicates as candidates and relied on automatic FrameNet-based semantic parsing and coarse lemma matching against article abstracts to identify salient propositions. \citet{choubey-etal-2020-discourse} annotated news articles for eight content type categories including `main events', but did not directly evaluate salience or produce graded scores. Also closely related to our work is the Pyramid method of \citet{nenkova-passonneau-2004-evaluating}, which evaluated generated summaries by aligning their content with semantic content units identified across multiple human reference summaries and weighted according to their frequency of occurrence. The method was developed using a subset of the Document Understanding Conference (DUC) data \citep{harman-over-2002-duc}. However they were not concerned with scoring salience as an end goal, and the data has not been released. Finally, \citet{zeldes-etal-2026-worth} approached gradient proposition salience evaluation directly and produced the test set we will use below for evaluation. But they provided no training or development data and did not attempt to automatically predict salience scores or evaluate computational models on the task in any way, a gap which this paper aims to close.

\subsection{Salience for extractive summarization}\label{sec:extractive}
The task of extractive summarization is closely related to 
proposition salience,
as it focuses on identifying which passages best describe a document's content. Early extractive summarization methods used sentence-similarity graphs to represent documents and unsupervised learning techniques to identify the most salient sentences \citep{mihalcea-tarau-2004-textrank, erkan-radev-2004-lexrank}. The advent of deep learning gave rise to neural network-based methods, primarily through supervised learning using pre-existing extractive summarization datasets \citep{cao-etal-2015-learning, cheng-lapata-2016-neural, nallapati-etal-2017-summa, ren-etal-2017-leverage, zhou-etal-2018-neural-document}. More recently, transformer architectures were used to identify salient sentences for summarization in specific target domains \citep{liu-lapata-2019-text, zhang-etal-2019-hibert}. Other approaches to salience combine encoder-based models with graph-based approaches that capture document position and inter-sentence redundancy. \citep{zheng-lapata-2019-sentence, jia-etal-2020-neural, wang-etal-2020-heterogeneous}. In addition, generative models have been used to produce extractive summaries as a sequence generation task but with mixed results \cite{zhang-etal-2023-extractive-summarization,le-luu-2024-extractive}. Finally, some work has also tested diffusion models on domain-specific extractive summarization \cite{dong-etal-2025-termdiffusum}. 

\subsection{RST-based approaches}

Rhetorical Structure Theory (RST) constructs hierarchical tree graphs over Elementary Discourse Units (EDUs) to express their relative importance (see Appendix \ref{sec:rst}).
Studies using RST to identify content for summarization found that RST tree nesting depth was a strong predictor of salience \citep{louis-etal-2010-discourse}. Further work showed that EDUs, instead of sentences, more closely matched content found in human-generated summaries, and that EDUs work as better units for near-extractive summarization \citep{li-etal-2016-role}. Later work leveraged EDUs as part of deep learning approaches to summarization for more concise text extraction \citep{liu-chen-2019-exploiting}. In DiscoBERT \citep{xu-etal-2020-discourse}, RST tree relationships were used to create Graph Convolutional Networks which showed significant improvement when compared with contemporary encoder models. A similar approach using a heterogeneous Graph model with EDUs performed comparably to DiscoBERT. \citet{xiao-etal-2020-really} modified an attention head in a BERT model with information about a document's RST tree and found improvements on extractive summarization using either EDUs or sentences as the minimal unit of analysis.  





\section{Data}

\begin{table*}[tbh]
\centering
\resizebox{\textwidth}{!}{%
\begin{tabular}{llrrrllrrr}
\toprule
\textbf{Genre} & \textbf{Source} & \textbf{Docs} & \textbf{EDUs} & \textbf{Alignments} & \textbf{Genre} & \textbf{Source} & \textbf{Docs} & \textbf{EDUs} & \textbf{Alignments}  \\
\midrule
Academic writing      & Various            & 18 & 1,981 & 9,905 & How-to guides       & wikiHow      & 19 & 2,399 & 11,995 \\
Biographies           & Wikipedia          & 20 &  2,071 & 10,355 & Interviews          & Wikinews     & 19 & 2,408 & 12,040 \\
Vlogs                 & YouTube   & 15 & 2,438 & 12,190 & Letters             & Various      & 18 & 2,102 & 10,510  \\
Conversations         & UCSB Corpus        & 15 & 3,083 & 15,415 & News stories        & Wikinews     & 24 & 1,912 & 9,560 \\
Courtroom transcripts & Various            & 14 & 2,143 & 10,715 & Podcasts            & Various      &  14 & 2,256 & 11,280 \\
Essays                & Various            & 14 & 2,001 & 10,005 & Political speeches  & Various      & 15 & 1,913 & 9,565 \\
Fiction               & CC fiction         & 19 & 2,474 & 12,370 & Textbooks           & OpenStax     & 15 & 2,030 & 10,150 \\
Forum discussions   & reddit             & 18 & 2,263 & 11,315 & Travel guides       & Wikivoyage   & 18 & 1,796 & 8,980 \\
\midrule
\textbf{Total docs / EDUs / aligns:} & \multicolumn{2}{l}{ 275 / 35,270 / 176,350} & \multicolumn{2}{l}{(dev: 32 / 3,896 / 19,480} & \multicolumn{2}{l}{test: 32 / 3,779 / 18,895 \quad\quad train:} & \multicolumn{3}{l}{ 211 / 27,595 / 137,975)} \\

\bottomrule
\end{tabular}
}
\caption{Overview of the \textsc{PropAlignBank} data from \textsc{GUM}.}
\label{tab:gum-data}
\end{table*}

\paragraph{Documents} As underlying text for \textsc{PropAlignBank} we use the freely available GUM (Georgetown University Multilayer corpus) data from \citet{zeldes-etal-2026-worth}, which provides five human written summaries for documents in the dev and test partitions of the corpus, as well as one human written and four model-generated summaries for the training set (see Table \ref{tab:gum-data}). The original data covers $\sim$290K tokens in over 35K propositions across 16 spoken and written genres of English, including face-to-face conversations, news and language on the Internet across modalities (e.g.~Reddit, YouTube vlogs, Wikipedia, and podcasts).

\paragraph{Manual annotation} \citet{zeldes-etal-2026-worth} provided manual alignments of EDUs to summaries in the GUM test set. For this study, we manually align all EDUs in the corresponding dev set to each of five summaries per document, resulting in $\sim$20K manual alignments, as shown in Table \ref{tab:gum-data}. Following the test set's manual, we label each alignment as \textcolor{orange}{\textit{good}}, \textcolor{cyan}{\textit{approximate}} (a likely trigger for content in the summary but not an exact match), or \textcolor{OliveGreen}{\textit{component}} (one part of content mentioned in the summary in aggregate), and track duplicate alignments (multiple propositions corresponding to or triggering the same summary text). As an example, consider the following summary and the text in \ref{ex:mangal-edus}. 

\medskip
\noindent Summary: ...\textit{editors of Global Voices discuss Mangal Shobhajatra, a festive  procession for the Bengali New Year that serves as a symbol of unity and diversity in Bangladesh}... \label{ex:mangal-summary}

\ex. $[$\textit{\textcolor{orange}{What you're hearing is Mangal Shobhajatra, an annual procession of thousands of people}}$]_1$ $[$\textit{\textcolor{cyan}{But what might look like just a street party is actually a creative stand for unity and against the forces of intolerance}}$]_2$ $[$\textit{\textcolor{OliveGreen}{I'm Lauren, news editor at Global Voices}}$]_3$  $[$\textit{\textcolor{OliveGreen}{I'm Sahar, managing editor at Global Voices}}$]_4$  $[$\textit{\textcolor{orange}{today we are digging deep into Bangladesh's Mangal Shobhajatra}}$]_5$ \label{ex:mangal-edus}

The units in \textcolor{orange}{orange} are considered good matches, the \textcolor{cyan}{cyan} unit is considered approximate (possible trigger for `symbol of unity and diversity'), and the \textcolor{OliveGreen}{green} units are components, since the summary mentions editors of Global Voices, aggregating over the two. Additionally, unit 5 is marked as a near duplicate of unit 1, since both align with `discuss Mangal Shobhajatra, an annual procession...'. We also include double annotations for 25  document-summary pairs and obtain inter-annotator agreement numbers (see Section \ref{sec:baseline-agreement}), which are in line with numbers from \citeauthor{zeldes-etal-2026-worth}. See Appendix \ref{sec:system prompts} for more detailed definitions which we provide to models.

\paragraph{Splits and additional annotations} Below we use the test split for scoring all algorithms, the dev split for few-shot examples and to train an ensemble approach leveraging multiple LLMs to predict on the test set. We use this ensemble to produce a high quality annotated version of the training set, for which summaries and EDU segments exist, but no alignments, with the future goal of enabling fine-tuning of open models on the task directly. Because the GUM data already includes a range of annotations, such as syntactic dependency trees, RST discourse parses and alignments of entities to summaries, we will leverage these annotations next to LLM alignment predictions in Section \ref{sec:ensemble} to improve prediction accuracy on the silver training set.

\section{Methods} 

\subsection{Settings}\label{sec:settings}

We evaluate several methods to extract salient propositions using LLMs:

\paragraph{\propdirect} As a baseline, we test model's direct sensitivity to proposition salience without providing any summaries. Because this baseline prompt does not reference a summary, we evaluate its outputs against summary-independent gold labels, which we construct by pooling human annotations across five summaries: An EDU is salient if it is annotated as salient for at least one summary. For a fair comparison, we apply the same pooling procedure to predictions from models using the summary-based approach, treating an EDU as salient if the model identifies it as salient for at least one summary.

\paragraph{\propaligned} For a finer-grained evaluation of the summary-based approach, we compare predictions against gold labels for each summary, testing whether models correctly identified that an EDU contributes to information in a specific summary. 

\paragraph{\propgraded} In this setting, we collect predictions from five summaries of each document and compute a single numeric score of 0--5 per proposition. We evaluate models purely on their ability to predict the correct number, and not whether those predictions are based on the correct summary (i.e.~a model can be right for the wrong reason if it answers positively for summary 1 and negatively for summary 2 when an EDU actually appeared in summary 2 but not 1).

\subsection{Models and prompts} 

While our preference would be to evaluate open-weight models, in practice we find that parameter efficient models severely underperform frontier models. We therefore prioritize comparisons of \mbox{GPT-5.6-terra} and \mbox{Claude-sonnet-5}, two recent, high-end but affordable models, and the smaller open-weight \mbox{Qwen-3.8-27B}. We first evaluate performance on \propdirect~by prompting single models, but store all responses with the goal of training a model ensemble fed by all predictions, which we evaluate below as well. We set reasoning to ``None'' for all API calls to keep costs manageable. Temperature is set to zero and Top K and Top P are not explicitly set.

To prompt models in the \propdirect~setting, we formulate a zero-shot, summary-free prompt which simply tasks models with identifying the most important units that describe the content of the document and a one-shot prompt with an example document and gold labels. The model is asked to output only the ID number of the salient units in a json format. See the full prompt in Appendix~\ref{sec:system prompts}.

In the other settings, models are given a summary and asked to identify units that contribute information to the summary in three steps: (1) attempt to align each EDU with a specific part of the given summary; (2) if successful, classify the alignment as \textit{good}, \textit{approximate}, or \textit{component}; (3) if the unit is redundant with a previously aligned unit for this summary, mark it as a duplicate of that ID. 
The prompt briefly defines each category and instructs models to include a short note on the corresponding text in the summary that prompted them to include it as aligned. See example outputs in Appendix~\ref{output} and  Appendix~\ref{sec:system prompts} for complete prompt templates. 

For prompt types, we compare a zero-shot setting and an in-genre one-shot setting, which uses a single document and summary pair to illustrate how the salience criteria apply to documents of the same genre (e.g.~if the model is predicting on a news article, the example is a different news article from the dev set). Note that this also applies to the \propdirect~ setting, in which the one-shot example can help the model to understand what is considered salient, even though here no summary is included. 

Since we have 10 dev set summary-document pairs for each genre, we make sure to choose a pair that contains all subtypes of match, including duplicate matches if possible. See the actual demonstration pairs and the number of subtypes included for each pair in Appendix~\ref{oneshot_demo}.

\subsection{Ensembles}

Because the train partition of our data is too large to annotate manually (\string~138K alignments), we generate the most accurate labels possible by ensembling predictions from LLMs together with additional annotations from the GUM dataset. As a metalearner algorithm, we test Random Forest and XGBoost, and predict alignments for each EDU-summary pair. 

\paragraph{Features} We utilize a combination of linguistic features and LLM-generated features. The LLM features come from prompting the Claude, GPT, and Qwen models and include both the binary judgments and predicted subtypes of each model. Linguistic features include lexical, positional, and similarity-based metrics, as well as annotations derived from the RST annotations available for GUM, such as RST relation name for each EDU and RST dependency depth, ROUGE scores for the summary vs. the EDU, cosine similarity based on sentence embeddings, and minimum frequency of any matching word between the summary and the EDU. A detailed list of features is given in Appendix \ref{sec:ens-features}.

\paragraph{Optimizing}
We tune each metalearner with 5-fold cross validation, with folds assigned at the document level. We utilize Optuna \citep{akiba2019optunanextgenerationhyperparameteroptimization} to optimize our Random Forest and XGBoost models. The search space covers model parameters, class imbalance handling, and the classification threshold (Appendix \ref{sec:app-ensemble}). For Random Forest, we perform backward elimination to prune out features. Features are ranked by decrease in impurity and are dropped if the cross-validated F1 score stays within a threshold of the best score seen so far. Full details for hyperparameter optimization are outlined in Appendix \ref{sec:app-ensemble}.

\subsection{Metrics}

The objective of the \textsc{PropAlign} method is to produce a graded salience score for each proposition in a text, which lends itself to reporting a ratio-scaled Absolute or Root Mean Squared Error (MAE/RMSE, how far predictions are from the target). However, in order to compare our method to naively prompting models to extract salient propositions, a binary setting may be more appropriate, where a positive judgment equates to `summary-worthiness' (i.e.~the proposition might appear in at least one summary). We refer to directly querying models for salience judgments as \propdirect~(Direct Proposition Salience), aligning propositions to a single summary as \propaligned~(Summary-aligned Proposition Salience), and aligning propositions to multiple summaries as \propgraded~(Graded-aligned Proposition Salience).

Since EDUs can be discontinuous (e.g.~interrupted by a relative clause, which is its own EDU), we evaluate model outputs on lists of IDs, where a positive judgment on any part of a discontinuous unit is considered to apply to the whole proposition. We report RMSE for \propgraded, as well as precision, recall, and F1 score for all settings, and we use mutual F1 score as a measure of agreement between models in both \propdirect~and the aligned settings, as well as to assess quality of our manually annotated data. Although we follow \citet{zeldes-etal-2026-worth} in tasking models with predicting whether alignments are good, approximate, or component alignments, we found that models do not reliably match human judgments and therefore use untyped judgments, ignoring these labels, as the deciding metric. We report subtype performance in Appendix~\ref{finegrainedeval}.

\section{Results} 
\begin{table}[t]
    \centering
    \setlength{\tabcolsep}{8pt}
    \renewcommand{\arraystretch}{1.15}
    \begin{tabular}{@{}lc@{}}
        \toprule
        Setting & F1 \\
        \midrule
        Baseline zero-shot & 0.464 \\
        Baseline one-shot & 0.638 \\
        Summary zero-shot & 0.716 \\
        Summary one-shot & 0.767 \\
        \bottomrule
    \end{tabular}
    \caption{Inter-model agreement between GPT and
    Claude, measured by F1 across experimental
    settings.}
    \label{tab:inter-model-agreement}
\end{table}
\subsection{\propdirect~baselines and agreement} \label{sec:baseline-agreement}

Table \ref{tab:inter-model-agreement} shows mutual F1 scores representing agreement between our two frontier models, \mbox{GPT-5.6-terra}  and \mbox{Claude-sonnet-5} for \propdirect~and \propaligned, the latter using a single summary (Qwen performance is substantially worse, see below).\footnote{Qwen sometimes does not return a cleanly parseable result as demanded by Appendix~\ref{output}, so we normalize the Qwen output for each summary by extracting all the identifiable numbers as the predicted salient unit ids. See Appendix~\ref{qwennormalization} for the number of documents that needed normalization.} We note first based on the table that adding a one-shot example to attune models to the targeted concept of salience is helpful (significantly better scores for one-shot over zero-shot, $p<0.05$). Second, we see a clear and significant improvement in agreement when using the summary based method, suggesting \textsc{PropAlign} is substantially more stable and reproducible than direct prompting.

Next we compare both zero-shot and one-shot scores against human reference annotations across the direct LLM baseline and the summary-based outputs. Table~\ref{tab:baseline-results} gives results under a pooled setting, where baseline performance uses no summary, and the summary-based comparison is pooled: if a model aligns an EDU to any of 5 summaries across 5 runs, the EDU is predicted to be `salient'. As can be seen from the table, summary-based approaches strongly outperform the baseline approaches, even when an example is provided. We perform McNemar's test between the best performing baseline one-shot model (Claude) and the worst performing summary-based zero-shot model (Qwen) and confirm that zero-shot summary-based Qwen shows a statistically significant improvement over baseline one-shot Claude, $\chi^2(1, N = 3{,}571) = 5.10$, $p = .024$. 

\begin{table}[t]
    \centering
    \setlength{\tabcolsep}{5pt}
    \renewcommand{\arraystretch}{1.15}
    \begin{tabular}{@{}llccc@{}}
        \toprule
        Setting & Model & P & R & F1 \\
        \midrule
        \multirow{3}{*}{\shortstack[l]{Baseline\\zero-shot}}
            & GPT & 0.742 & 0.161 & 0.264 \\
            & Claude & 0.594 & 0.332 & 0.426 \\
            & Qwen & 0.720 & 0.089 & 0.159 \\
        \midrule
        \multirow{3}{*}{\shortstack[l]{Baseline\\one-shot}}
            & GPT & 0.548 & 0.451 & 0.495 \\
            & Claude & 0.517 & 0.517 & 0.517 \\
            & Qwen & 0.575 & 0.293 & 0.388 \\
        \midrule
        \multirow{3}{*}{\shortstack[l]{Summary\\zero-shot}}
            & GPT & 0.509 & 0.850 & 0.637 \\
            & Claude & 0.650 & 0.728 & 0.687 \\
            & Qwen & 0.555 & 0.606 & 0.580 \\
        \midrule
        \multirow{3}{*}{\shortstack[l]{Summary\\one-shot}}
            & GPT & 0.606 & 0.802 & 0.690 \\
            & Claude & 0.680 & 0.739 & 0.709 \\
            & Qwen & 0.540 & 0.696 & 0.608 \\
        \midrule
        Humans & -- & 0.708 & 0.698 & 0.703 \\
        \bottomrule
    \end{tabular}
    \caption{Precision (P), recall (R), and F1 across
    models and experimental settings, evaluated against
    aggregated human gold labels and output IDs.}
    \label{tab:baseline-results}
    \vspace{-1em}
\end{table}

The summary-based method allows the much smaller Qwen model to become competitive with frontier models -- one-shot summary-based Qwen outperforms baseline one-shot GPT and Claude without the summary-based task setup. The gap between Qwen and the frontier models in the summary based  experiments is substantially smaller too (about 10 points, compared to the 27 point gap between zero shot baseline Qwen and the best frontier model in that setup, Claude). This suggests that the more well-defined and stable task setup can enable the use of much more cost-effective models for salient proposition extraction.

The table also reports similarly computed mutual P/R/F1 using double annotated data, evaluated in a binary setting (salient for any summary/non-salient), to match model scoring in this setting. Here we note that summary-based one-shot Claude approaches the human ceiling under this setting, showing no statistically significant difference, though humans do actually outperform all LLMs on EDU-summary aligned pairs (see \propgraded~results below).



\subsection{\propaligned~performance}

Next, we ask each model to align EDUs to the existing summaries and score using fine-grained evaluation, based on precision and recall of each EDU-summary alignment pair (non-pooled micro-average). Table~\ref{tab:main-results} shows that in-genre one-shot demonstration helps for GPT and Qwen significantly ($p<0.05$) in this setting, but surprisingly not for Claude. Using McNemar's test with continuity correction, there is no statistical difference between the outputs of Claude zero-shot and Claude one-shot, $\chi^2(1, N = 17{,}855) = 0.02$, $p = .897$. Although Claude is the best single model for obtaining pairwise alignments for a \propaligned~approach, other models are competitive in obtaining the final graded scores across all five summaries, which are much more stable. We turn to these next.

\begin{table}[t]
    \centering
    \setlength{\tabcolsep}{5pt}
    \renewcommand{\arraystretch}{1.2}
    \resizebox{\columnwidth}{!}{%
    \begin{tabular}{@{}llccccc@{}}
        \toprule
        Setting & Model & P & R & F1 & MAE & RMSE \\
        \midrule
        \multirow{3}{*}{\shortstack[l]{Summary\\zero-shot}}
            & GPT & 0.401 & 0.717 & 0.515 & 0.755 & 1.38 \\
        \cmidrule(l){2-7}
            & Claude & 0.568 & 0.639 & 0.601 & 0.466 & 0.996 \\
        \cmidrule(l){2-7}
            & Qwen & 0.507 & 0.392 & 0.431 & 0.533 & 1.035 \\
        \midrule
        \multirow{3}{*}{\shortstack[l]{Summary\\one-shot}}
            & GPT & 0.480 & 0.679 & 0.563 & 0.583 & 1.19 \\
        \cmidrule(l){2-7}
            & Claude & 0.572 & 0.629 & 0.599 & 0.452 & 0.993 \\
        \cmidrule(l){2-7}
            & Qwen & 0.478 & 0.504 & 0.496 & 0.515 & 0.973 \\
        \midrule
        Ensemble & & 0.592 & 0.655 & 0.622 & 0.431 & 0.988 \\
        \midrule
        Humans & & 0.708 & 0.698 & 0.703 & 0.328 & 0.798 \\
        \bottomrule
    \end{tabular}
    }
    \caption{Precision (P), recall (R), F1, mean absolute error
    (MAE), and root mean squared error (RMSE) across experimental
    settings, the ensemble, and human performance for the fine-grained
    evaluation scheme.} 
    \label{tab:main-results}
    \vspace{-1em}
\end{table}

\subsection{\propgraded~and Ensemble} \label{sec:ensemble}

Switching from mutual P/R/F1 scores to numeric estimates of deviation from the target score of each EDU (0--5), we again see better performance for one-shot predictions in each model (lower MAE and RMSE). Claude scores the best in MAE, consistent with its performance in F1. Although Qwen achieves the lowest overall RMSE, this is largely because its errors consist primarily of small deviations from the gold salience scores, which RMSE penalizes less heavily than Claude's less frequent but larger errors. This contrasts with Qwen's poor F1 score. These seemingly conflicting facts stem from the fact that in our summary one-shot experiment, Qwen often obtained a correct total salience score by relying on incorrect individual alignments (false positives compensating for false negatives); in other words, it is sometimes right for the wrong reasons, or often has deviations of just 1 point from 0 -- see details in Appendix~\ref{saliencescore}. However a 1 point deviation above 0 (somewhat salient vs. not at all) may be more significant for downstream users than a numerically larger error between predictions of 2 and 4, which penalize especially Claude.

To construct the best possible predictions for the full \textsc{PropAlignBank} dataset, we found that including the binary salience predictions from GPT and Claude in a tuned Random Forest, in addition to the linguistic features, gives the best F1 score.  In addition, during model development, we found that the LLM-generated labels for duplicate, approximate, and component propositions did not improve performance, and therefore we retained only the binary salience predictions as our LLM-based features.  This ensemble method also gives the best MAE score. Although this method is not usable in realistic settings for unseen data, since the ensemble leverages manual annotations from the GUM corpus, we are able to use it to annotate the training partition automatically, which we release as an additional silver-standard partition of our dataset.

We perform McNemar's test with continuity correction on the difference between the best performing single model and the ensemble. The tuned Random Forest ensemble model shows a statistically significant improvement over Claude zero-shot and Claude one-shot,
$\chi^2(1, N = 17{,}855) = 20.59$, $p < .001$ and $\chi^2(1, N = 17{,}855) = 14.61$, $p < .001$ respectively. In addition, human performance still shows a statistically significant
advantage over the ensemble ($\chi^2(1, N = 8{,}585) = 39.47$, $p < .001$) respectively, which is our best model, suggesting room
for improvement on the task.\footnote{For half of the test set, two annotators independently annotated the data. The gold annotations were adjudicated by \citet{zeldes-etal-2026-worth}. However, when evaluating our human performance, we compute mutual agreement on each annotator's judgments, without adjudication.} 


\section{Error analysis}
\subsection{Quantitative analysis}
\begin{figure*}
\centering
  \includegraphics[width=1.0\textwidth]{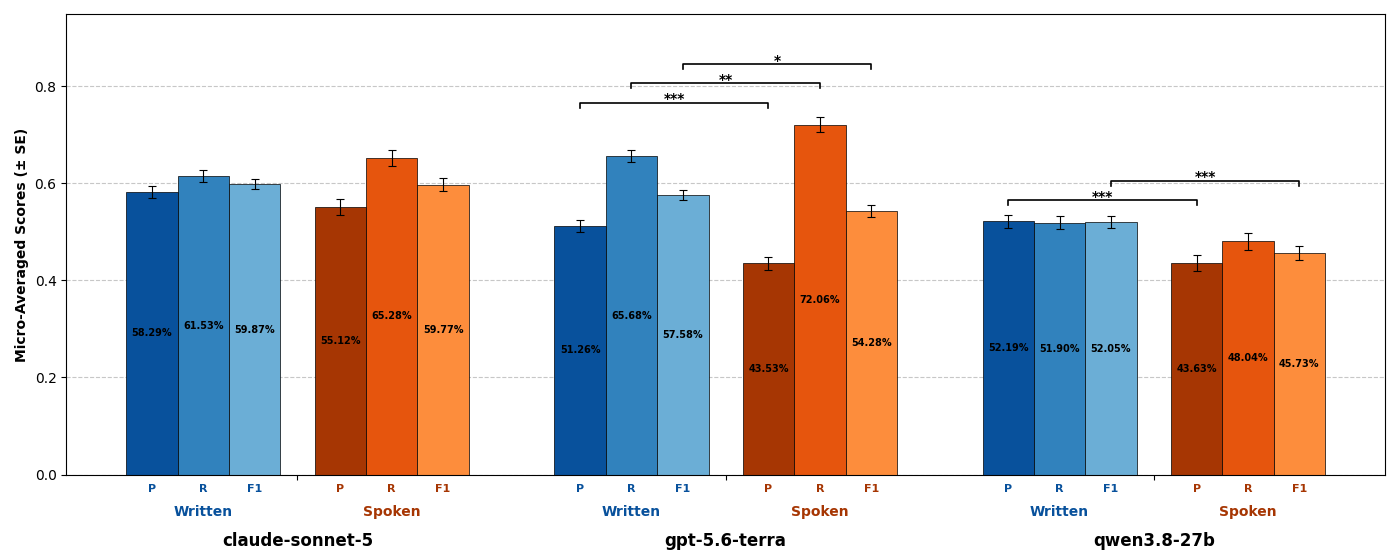}
  \caption{Precision, recall and F1 for three models on written and spoken test data.}
  \label{fig:modality-prf}
  \vspace{-6pt}
\end{figure*}

Figure \ref{fig:modality-prf} shows model performance on spoken versus written data. The best model, Claude, shows no significant difference across modalities, though precision is lower than recall. GPT and Qwen show significantly better F1 and precision on written data, though GPT's recall is significantly better on the spoken modality. Low precision on speech is the weakest point for all three models. 

There were also effects noted on the position of EDUs in the document, as errors were most concentrated in the first 30\% of EDUs across all models. The effects were most noticeable with Claude and Qwen having over 15\% of errors in the first 10\% of EDUs. A more detailed analysis of error types and position is provided in Appendix \ref{sec:error-position}.

\subsection{Qualitative analysis}

Looking at the lowest-scoring documents for the frontier models in the development set, we observe multiple failures in identification of levels of granularity and generalization. For example, in analyzing a Reddit post about macroeconomics discussing how and to whom countries can be in debt, summary-one shot GPT flags the following false overlaps with the summary, which humans skipped:

\medskip
\noindent Summary: \textit{...author explains how money is a form of debt whose value depends on the amount of money in circulation ...} 

\ex. \textit{[You can borrow from your own people]} 

\ex. \textit{[if the govt decided to print 1 quadrillion dollars,] [the value of the dollar would probably crash pretty hard]}\label{ex:macroecon}

Although the summary does not mention borrowing from a government's people, the notions of borrowing and debt are closely related, and the model may generalize from phrases such as `amount of money in circulation' to a reference to the people in a country. In the second error, devaluation of currency is related to value depending on the amount of money in circulation, but the summary does not discuss the cited example involving the US dollar, and there are better, more general trigger units in the document to align to `value depends on the amount of money in circulation'.

Conversely, for a conversation during a game of Risk, the summary mentions that the interlocutors are playing with cards. The gold annotations flag EDUs that specifically reference the users holding cards as \textcolor{cyan}{approximate} matches:

\medskip

\noindent Summary: \textit{Two people are playing a strategy game online involving cards...} 

\ex. \textit{[\textcolor{cyan}{How many cards do you have?}]} \label{ex:risk1} 

\ex. \textit{[\textcolor{cyan}{You get the cards.}]} \label{ex:risk2} 

In both \ref{ex:risk1} and \ref{ex:risk2}, the human annotators utilize the mention of cards as a trigger for the content in the summary, which lacks other sources. However, Claude does not highlight either of these EDUs, instead opting for a mention of the word \textit{set} (which indirectly references the cards) as in \ref{ex:risk3}, despite the presence of obviously more related propositions in the document.

\ex. \textit{[So you can't have a set.]} \label{ex:risk3}

\section{Discussion}

\paragraph{Why should we use summary based salience?} Given the results, it is clear that naively asking models to extract salient information is substantially less stable than \textsc{PropAlign}. This result may seem unsurprising: if models  don't know that we define salience based on alignment with a summary that they are not provided, they will make mistakes. However, note that the same applies to direct prompting: without a clear operationalization, models are bound to disagree, and this study shows that agreement levels are much better for \textsc{PropAlign}, which is non-trivial. Additionally, using multiple summaries as components of a gradient salience score not only results in higher convergence across models, but also offers a more explainable criterion for quantitative salience ratings.

A second reason to seriously consider using \textsc{PropAlign} instead of direct querying methods is the very strong improvement in performance for the smaller Qwen model even in the zero-shot \textsc{SaPSal} scenario -- the simplest possible summary-based methodology. In this scenario, the much smaller Qwen model lags behind Claude Sonnet-5 by only 10 points, compared to the 27 point gap in \textsc{DPSal}. The jump to the best \textsc{GaPSal} method allows Qwen to go neck-and-neck with Claude in RMSE for gradient scores, 
though not in F1 score, and Claude has the advantage in MAE as well.

\paragraph{Isn't this too complex to be practical?} In production settings, the pipeline evaluated in this paper would involve obtaining multiple summaries per document and then aligning them to documents using multiple model runs, resulting in an expensive and complex setup. While this is indisputable, we emphasize that naive alternatives are ill defined, as seen from the lower scores in the baseline setting in Table \ref{tab:baseline-results} and the lower level of inter-model agreement in the baseline setting in Table \ref{tab:inter-model-agreement}. 

That being said, the silver-standard training set created in this paper, with around 138K alignments, offers the potential to make further progress. The data could be used to study what makes propositions salient in discourse more directly, as well as to train systems on the more consistent concept of salience promoted here in order to predict it with a single fine-tuned model prompt, a possibility which we leave for further study.

\section*{Limitations}

This paper presupposes, rather than proves, that summary-based salience is a desirable target for quantifying salience, and does not compare the performance of its methods on possibly related pre-existing tasks, such as extractive summarization. While we acknowledge an interesting relationship between salience metrics and extractive summarization, we consider these conceptually separate tasks: extractive summarization may have different priorities in specific domains, and is also concerned with intelligibility of extracts (for example problems created by the need to reconstruct pronouns, or insufficient context), whereas our method is fully agnostic of intelligibility in isolation. Similarly, approximate and component matches discussed above may make for poor extractive summaries, but we consider them salient since they are required triggers for content in the abstractive summaries used by our method.

Although our study covers a broad sample of 16 spoken and written genres, it is limited to English language sources. We are optimistic that our methods will be easy to apply to data in other languages, our results especially with regard to LLM performance may or may not generalize. We also recognize that a full application of the pipeline including collection of five abstractive summaries and alignment is relatively expensive compared to a \propdirect~approach, and we leave it to future studies to leverage \textsc{PropAlignBank} to fine-tune models end-to-end on the task data.

All annotations collected in this study were carried out by Computational Linguistics graduate students or post-grads on a voluntary basis as part of an academic research project. LLMs were accessed via the Open AI and OpenRouter APIs, at a total cost of \$35.



\bibliography{custom}

@inproceedings{harman-over-2002-duc,
  author    = {Harman, Donna K. and Over, Paul D.},
  title     = {The {DUC} Summarization Evaluations},
  booktitle = {Proceedings of the Second International Conference on Human Language Technology Research},
  year      = {2002},
}

@inproceedings{zeldes-etal-2026-worth,
    title = "Not Worth Mentioning? {A} Pilot Study on Salient Proposition Annotation",
    author = "Zeldes, Amir  and
      Conhaim, Katherine  and
      Levine, Lauren",
    editor = "Liu, Yang Janet  and
      Gessler, Luke",
    booktitle = "Proceedings of the 20th Linguistic Annotation Workshop ({LAW} {XX})",
    month = jul,
    year = "2026",
    address = "San Diego, California, USA",
    publisher = "Association for Computational Linguistics",
    url = "https://aclanthology.org/2026.law-main.14/",
    doi = "10.18653/v1/2026.law-main.14",
    pages = "178--186",
    ISBN = "979-8-89176-404-0"
}

@misc{akiba2019optunanextgenerationhyperparameteroptimization,
      title={Optuna: A Next-generation Hyperparameter Optimization Framework}, 
      author={Takuya Akiba and Shotaro Sano and Toshihiko Yanase and Takeru Ohta and Masanori Koyama},
      year={2019},
      eprint={1907.10902},
      archivePrefix={arXiv},
      primaryClass={cs.LG},
      url={https://arxiv.org/abs/1907.10902}, 
}

@inproceedings{dunietz-gillick-2014-new,
    title = "A New Entity Salience Task with Millions of Training Examples",
    author = "Dunietz, Jesse  and
      Gillick, Daniel",
    editor = "Wintner, Shuly  and
      Riezler, Stefan  and
      Goldwater, Sharon",
    booktitle = "Proceedings of the 14th Conference of the {E}uropean Chapter of the Association for Computational Linguistics, volume 2: Short Papers",
    month = apr,
    year = "2014",
    address = "Gothenburg, Sweden",
    publisher = "Association for Computational Linguistics",
    url = "https://aclanthology.org/E14-4040/",
    doi = "10.3115/v1/E14-4040",
    pages = "205--209"
}

@article{erkan-radev-2004-lexrank,
author = {Erkan, G{\"u}nes and Radev, Dragomir R.},
title = {LexRank: graph-based lexical centrality as salience in text summarization},
year = {2004},
issue_date = {July 2004},
publisher = {AI Access Foundation},
address = {El Segundo, CA, USA},
volume = {22},
number = {1},
issn = {1076-9757},
journal = {J. Artif. Int. Res.},
month = dec,
pages = {457–479},
numpages = {23}
}

@inproceedings{mihalcea-tarau-2004-textrank,
    title = "{T}ext{R}ank: Bringing Order into Text",
    author = "Mihalcea, Rada  and
      Tarau, Paul",
    editor = "Lin, Dekang  and
      Wu, Dekai",
    booktitle = "Proceedings of the 2004 Conference on Empirical Methods in Natural Language Processing",
    month = jul,
    year = "2004",
    address = "Barcelona, Spain",
    publisher = "Association for Computational Linguistics",
    url = "https://aclanthology.org/W04-3252/",
    pages = "404--411"
}

@inproceedings{cheng-lapata-2016-neural,
    title = "Neural Summarization by Extracting Sentences and Words",
    author = "Cheng, Jianpeng  and
      Lapata, Mirella",
    editor = "Erk, Katrin  and
      Smith, Noah A.",
    booktitle = "Proceedings of the 54th Annual Meeting of the Association for Computational Linguistics (Volume 1: Long Papers)",
    month = aug,
    year = "2016",
    address = "Berlin, Germany",
    publisher = "Association for Computational Linguistics",
    url = "https://aclanthology.org/P16-1046/",
    doi = "10.18653/v1/P16-1046",
    pages = "484--494"
}

@inproceedings{nallapati-etal-2017-summa,
author = {Nallapati, Ramesh and Zhai, Feifei and Zhou, Bowen},
title = {SummaRuNNer: a recurrent neural network based sequence model for extractive summarization of documents},
year = {2017},
publisher = {AAAI Press},
booktitle = {Proceedings of the Thirty-First AAAI Conference on Artificial Intelligence},
pages = {3075–3081},
numpages = {7},
location = {San Francisco, California, USA},
series = {AAAI'17}
}

@inproceedings{zhou-etal-2018-neural-document,
    title = "Neural Document Summarization by Jointly Learning to Score and Select Sentences",
    author = "Zhou, Qingyu  and
      Yang, Nan  and
      Wei, Furu  and
      Huang, Shaohan  and
      Zhou, Ming  and
      Zhao, Tiejun",
    editor = "Gurevych, Iryna  and
      Miyao, Yusuke",
    booktitle = "Proceedings of the 56th Annual Meeting of the Association for Computational Linguistics (Volume 1: Long Papers)",
    month = jul,
    year = "2018",
    address = "Melbourne, Australia",
    publisher = "Association for Computational Linguistics",
    url = "https://aclanthology.org/P18-1061/",
    doi = "10.18653/v1/P18-1061",
    pages = "654--663"
}

@inproceedings{cao-etal-2015-learning,
    title = "Learning Summary Prior Representation for Extractive Summarization",
    author = "Cao, Ziqiang  and
      Wei, Furu  and
      Li, Sujian  and
      Li, Wenjie  and
      Zhou, Ming  and
      Wang, Houfeng",
    editor = "Zong, Chengqing  and
      Strube, Michael",
    booktitle = "Proceedings of the 53rd Annual Meeting of the Association for Computational Linguistics and the 7th International Joint Conference on Natural Language Processing (Volume 2: Short Papers)",
    month = jul,
    year = "2015",
    address = "Beijing, China",
    publisher = "Association for Computational Linguistics",
    url = "https://aclanthology.org/P15-2136/",
    doi = "10.3115/v1/P15-2136",
    pages = "829--833"
}

@inproceedings{ren-etal-2017-leverage,
    author = {Ren, Pengjie and Chen, Zhumin and Ren, Zhaochun and Wei, Furu and Ma, Jun and de Rijke, Maarten},
    title = {Leveraging Contextual Sentence Relations for Extractive Summarization Using a Neural Attention Model},
    year = {2017},
    isbn = {9781450350228},
    publisher = {Association for Computing Machinery},
    address = {New York, NY, USA},
    url = {https://doi.org/10.1145/3077136.3080792},
    doi = {10.1145/3077136.3080792},
    booktitle = {Proceedings of the 40th International ACM SIGIR Conference on Research and Development in Information Retrieval},
    pages = {95–104},
    numpages = {10},
    location = {Shinjuku, Tokyo, Japan},
    series = {SIGIR '17}
    }

@inproceedings{liu-lapata-2019-text,
    title = "Text Summarization with Pretrained Encoders",
    author = "Liu, Yang  and
      Lapata, Mirella",
    editor = "Inui, Kentaro  and
      Jiang, Jing  and
      Ng, Vincent  and
      Wan, Xiaojun",
    booktitle = "Proceedings of the 2019 Conference on Empirical Methods in Natural Language Processing and the 9th International Joint Conference on Natural Language Processing (EMNLP-IJCNLP)",
    month = nov,
    year = "2019",
    address = "Hong Kong, China",
    publisher = "Association for Computational Linguistics",
    url = "https://aclanthology.org/D19-1387/",
    doi = "10.18653/v1/D19-1387",
    pages = "3730--3740"
}

@inproceedings{zheng-lapata-2019-sentence,
    title = "Sentence Centrality Revisited for Unsupervised Summarization",
    author = "Zheng, Hao  and
      Lapata, Mirella",
    editor = "Korhonen, Anna  and
      Traum, David  and
      M{\`a}rquez, Llu{\'i}s",
    booktitle = "Proceedings of the 57th Annual Meeting of the Association for Computational Linguistics",
    month = jul,
    year = "2019",
    address = "Florence, Italy",
    publisher = "Association for Computational Linguistics",
    url = "https://aclanthology.org/P19-1628/",
    doi = "10.18653/v1/P19-1628",
    pages = "6236--6247"
}

@inproceedings{zhang-etal-2019-hibert,
    title = "{HIBERT}: Document Level Pre-training of Hierarchical Bidirectional Transformers for Document Summarization",
    author = "Zhang, Xingxing  and
      Wei, Furu  and
      Zhou, Ming",
    editor = "Korhonen, Anna  and
      Traum, David  and
      M{\`a}rquez, Llu{\'i}s",
    booktitle = "Proceedings of the 57th Annual Meeting of the Association for Computational Linguistics",
    month = jul,
    year = "2019",
    address = "Florence, Italy",
    publisher = "Association for Computational Linguistics",
    url = "https://aclanthology.org/P19-1499/",
    doi = "10.18653/v1/P19-1499",
    pages = "5059--5069"
}

@inproceedings{wang-etal-2020-heterogeneous,
    title = "Heterogeneous Graph Neural Networks for Extractive Document Summarization",
    author = "Wang, Danqing  and
      Liu, Pengfei  and
      Zheng, Yining  and
      Qiu, Xipeng  and
      Huang, Xuanjing",
    editor = "Jurafsky, Dan  and
      Chai, Joyce  and
      Schluter, Natalie  and
      Tetreault, Joel",
    booktitle = "Proceedings of the 58th Annual Meeting of the Association for Computational Linguistics",
    month = jul,
    year = "2020",
    address = "Online",
    publisher = "Association for Computational Linguistics",
    url = "https://aclanthology.org/2020.acl-main.553/",
    doi = "10.18653/v1/2020.acl-main.553",
    pages = "6209--6219"
}

@Article{MannThompson1988,
  title                    = {{Rhetorical Structure Theory}: Toward a Functional Theory of Text Organization},
  author                   = {William C. Mann and Sandra A. Thompson},
  journal                  = {Text},
  year                     = {1988},
  number                   = {3},
  pages                    = {243-281},
  volume                   = {8},
}

@inproceedings{jia-etal-2020-neural,
    title = "Neural Extractive Summarization with Hierarchical Attentive Heterogeneous Graph Network",
    author = "Jia, Ruipeng  and
      Cao, Yanan  and
      Tang, Hengzhu  and
      Fang, Fang  and
      Cao, Cong  and
      Wang, Shi",
    editor = "Webber, Bonnie  and
      Cohn, Trevor  and
      He, Yulan  and
      Liu, Yang",
    booktitle = "Proceedings of the 2020 Conference on Empirical Methods in Natural Language Processing (EMNLP)",
    month = nov,
    year = "2020",
    address = "Online",
    publisher = "Association for Computational Linguistics",
    url = "https://aclanthology.org/2020.emnlp-main.295/",
    doi = "10.18653/v1/2020.emnlp-main.295",
    pages = "3622--3631"
}

@inproceedings{liu-chen-2019-exploiting,
    title = "Exploiting Discourse-Level Segmentation for Extractive Summarization",
    author = "Liu, Zhengyuan  and
      Chen, Nancy",
    editor = "Wang, Lu  and
      Cheung, Jackie Chi Kit  and
      Carenini, Giuseppe  and
      Liu, Fei",
    booktitle = "Proceedings of the 2nd Workshop on New Frontiers in Summarization",
    month = nov,
    year = "2019",
    address = "Hong Kong, China",
    publisher = "Association for Computational Linguistics",
    url = "https://aclanthology.org/D19-5415/",
    doi = "10.18653/v1/D19-5415",
    pages = "116--121"
}

@inproceedings{xu-etal-2020-discourse,
    title = "Discourse-Aware Neural Extractive Text Summarization",
    author = "Xu, Jiacheng  and
      Gan, Zhe  and
      Cheng, Yu  and
      Liu, Jingjing",
    editor = "Jurafsky, Dan  and
      Chai, Joyce  and
      Schluter, Natalie  and
      Tetreault, Joel",
    booktitle = "Proceedings of the 58th Annual Meeting of the Association for Computational Linguistics",
    month = jul,
    year = "2020",
    address = "Online",
    publisher = "Association for Computational Linguistics",
    url = "https://aclanthology.org/2020.acl-main.451/",
    doi = "10.18653/v1/2020.acl-main.451",
    pages = "5021--5031"
}

@inproceedings{xiao-etal-2020-really,
    title = "Do We Really Need That Many Parameters In Transformer For Extractive Summarization? Discourse Can Help !",
    author = "Xiao, Wen  and
      Huber, Patrick  and
      Carenini, Giuseppe",
    editor = "Braud, Chlo{\'e}  and
      Hardmeier, Christian  and
      Li, Junyi Jessy  and
      Louis, Annie  and
      Strube, Michael",
    booktitle = "Proceedings of the First Workshop on Computational Approaches to Discourse",
    month = nov,
    year = "2020",
    address = "Online",
    publisher = "Association for Computational Linguistics",
    url = "https://aclanthology.org/2020.codi-1.13/",
    doi = "10.18653/v1/2020.codi-1.13",
    pages = "124--134"
}

@inproceedings{wu-etal-2023-edu,
    title = "{EDU}-level Extractive Summarization with Varying Summary Lengths",
    author = "Wu, Yuping  and
      Tseng, Ching-Hsun  and
      Shang, Jiayu  and
      Mao, Shengzhong  and
      Nenadic, Goran  and
      Zeng, Xiao-Jun",
    editor = "Vlachos, Andreas  and
      Augenstein, Isabelle",
    booktitle = "Findings of the Association for Computational Linguistics: EACL 2023",
    month = may,
    year = "2023",
    address = "Dubrovnik, Croatia",
    publisher = "Association for Computational Linguistics",
    url = "https://aclanthology.org/2023.findings-eacl.123/",
    doi = "10.18653/v1/2023.findings-eacl.123",
    pages = "1655--1667"
}

@inproceedings{louis-etal-2010-discourse,
    title = "Discourse indicators for content selection in summarization",
    author = "Louis, Annie  and
      Joshi, Aravind  and
      Nenkova, Ani",
    editor = "Katagiri, Yasuhiro  and
      Nakano, Mikio",
    booktitle = "Proceedings of the {SIGDIAL} 2010 Conference",
    month = sep,
    year = "2010",
    address = "Tokyo, Japan",
    publisher = "Association for Computational Linguistics",
    url = "https://aclanthology.org/W10-4327/",
    pages = "147--156"
}

@inproceedings{li-etal-2016-role,
    title = "The Role of Discourse Units in Near-Extractive Summarization",
    author = "Li, Junyi Jessy  and
      Thadani, Kapil  and
      Stent, Amanda",
    editor = "Fernandez, Raquel  and
      Minker, Wolfgang  and
      Carenini, Giuseppe  and
      Higashinaka, Ryuichiro  and
      Artstein, Ron  and
      Gainer, Alesia",
    booktitle = "Proceedings of the 17th Annual Meeting of the Special Interest Group on Discourse and Dialogue",
    month = sep,
    year = "2016",
    address = "Los Angeles",
    publisher = "Association for Computational Linguistics",
    url = "https://aclanthology.org/W16-3617/",
    doi = "10.18653/v1/W16-3617",
    pages = "137--147"
}

@inproceedings{le-luu-2024-extractive,
    title = "Extractive Summarization with Text Generator",
    author = "Le, Thang  and
      Luu, Anh Tuan",
    editor = "Duh, Kevin  and
      Gomez, Helena  and
      Bethard, Steven",
    booktitle = "Proceedings of the 2024 Conference of the North American Chapter of the Association for Computational Linguistics: Human Language Technologies (Volume 1: Long Papers)",
    month = jun,
    year = "2024",
    address = "Mexico City, Mexico",
    publisher = "Association for Computational Linguistics",
    url = "https://aclanthology.org/2024.naacl-long.9/",
    doi = "10.18653/v1/2024.naacl-long.9",
    pages = "157--174"
}

@inproceedings{zhang-etal-2023-extractive-summarization,
    title = "Extractive Summarization via {C}hat{GPT} for Faithful Summary Generation",
    author = "Zhang, Haopeng  and
      Liu, Xiao  and
      Zhang, Jiawei",
    editor = "Bouamor, Houda  and
      Pino, Juan  and
      Bali, Kalika",
    booktitle = "Findings of the Association for Computational Linguistics: EMNLP 2023",
    month = dec,
    year = "2023",
    address = "Singapore",
    publisher = "Association for Computational Linguistics",
    url = "https://aclanthology.org/2023.findings-emnlp.214/",
    doi = "10.18653/v1/2023.findings-emnlp.214",
    pages = "3270--3278"
}

@inproceedings{dong-etal-2025-termdiffusum,
    title = "{T}erm{D}iffu{S}um: A Term-guided Diffusion Model for Extractive Summarization of Legal Documents",
    author = "Dong, Xiangyun  and
      Li, Wei  and
      Le, Yuquan  and
      Jiang, Zhangyue  and
      Zhong, Junxi  and
      Wang, Zhong",
    editor = "Rambow, Owen  and
      Wanner, Leo  and
      Apidianaki, Marianna  and
      Al-Khalifa, Hend  and
      Eugenio, Barbara Di  and
      Schockaert, Steven",
    booktitle = "Proceedings of the 31st International Conference on Computational Linguistics",
    month = jan,
    year = "2025",
    address = "Abu Dhabi, UAE",
    publisher = "Association for Computational Linguistics",
    url = "https://aclanthology.org/2025.coling-main.216/",
    pages = "3222--3235"
}

@article{PengGeWang2023,
  author  = {Cheng Peng and Tong Ge and Yunhai Wang},
  title   = {Optimizing Word Search Within Documents by Showing Results in the Context},
  journal = {Journal of Visualization},
  volume  = {26},
  number  = {2},
  pages   = {441--455},
  year    = {2023},
  doi     = {10.1007/s12650-022-00880-5}
}

@inproceedings{zhou-etal-2020-level,
    title = "At Which Level Should We Extract? An Empirical Analysis on Extractive Document Summarization",
    author = "Zhou, Qingyu  and
      Wei, Furu  and
      Zhou, Ming",
    editor = "Scott, Donia  and
      Bel, Nuria  and
      Zong, Chengqing",
    booktitle = "Proceedings of the 28th International Conference on Computational Linguistics",
    month = dec,
    year = "2020",
    address = "Barcelona, Spain (Online)",
    publisher = "International Committee on Computational Linguistics",
    url = "https://aclanthology.org/2020.coling-main.492/",
    doi = "10.18653/v1/2020.coling-main.492",
    pages = "5617--5628"
}

@techreport{CarlsonMarcu2001,
  author      = {Lynn Carlson and Daniel Marcu},
  title       = {Discourse Tagging Reference Manual},
  institution = {Information Sciences Institute, University of Southern California},
  type        = {Technical Report},
  year        = {2001},
  month       = sep
}

@inproceedings{bar-haim-etal-2020-quantitative,
    title = "Quantitative argument summarization and beyond: Cross-domain key point analysis",
    author = "Bar-Haim, Roy  and
      Kantor, Yoav  and
      Eden, Lilach  and
      Friedman, Roni  and
      Lahav, Dan  and
      Slonim, Noam",
    editor = "Webber, Bonnie  and
      Cohn, Trevor  and
      He, Yulan  and
      Liu, Yang",
    booktitle = "Proceedings of the 2020 Conference on Empirical Methods in Natural Language Processing (EMNLP)",
    month = nov,
    year = "2020",
    address = "Online",
    publisher = "Association for Computational Linguistics",
    url = "https://aclanthology.org/2020.emnlp-main.3/",
    doi = "10.18653/v1/2020.emnlp-main.3",
    pages = "39--49"
}

@inproceedings{bullough-etal-2024-predicting,
    title = "Predicting Entity Salience in Extremely Short Documents",
    author = "Bullough, Benjamin  and
      Lundberg, Harrison  and
      Hu, Chen  and
      Xiao, Weihang",
    editor = "Dernoncourt, Franck  and
      Preo{\c{t}}iuc-Pietro, Daniel  and
      Shimorina, Anastasia",
    booktitle = "Proceedings of the 2024 Conference on Empirical Methods in Natural Language Processing: Industry Track",
    month = nov,
    year = "2024",
    address = "Miami, Florida, US",
    publisher = "Association for Computational Linguistics",
    url = "https://aclanthology.org/2024.emnlp-industry.5/",
    doi = "10.18653/v1/2024.emnlp-industry.5",
    pages = "50--64"
}

@inproceedings{lin-zeldes-2024-gumsley,
    title = "{GUM}sley: Evaluating Entity Salience in Summarization for 12 {E}nglish Genres",
    author = "Lin, Jessica  and
      Zeldes, Amir",
    editor = "Graham, Yvette  and
      Purver, Matthew",
    booktitle = "Proceedings of the 18th Conference of the European Chapter of the Association for Computational Linguistics (Volume 1: Long Papers)",
    month = mar,
    year = "2024",
    address = "St. Julian{'}s, Malta",
    publisher = "Association for Computational Linguistics",
    url = "https://aclanthology.org/2024.eacl-long.158/",
    doi = "10.18653/v1/2024.eacl-long.158",
    pages = "2575--2588"
}

@inproceedings{choubey-etal-2020-discourse,
    title = "Discourse as a Function of Event: Profiling Discourse Structure in News Articles around the Main Event",
    author = "Choubey, Prafulla Kumar  and
      Lee, Aaron  and
      Huang, Ruihong  and
      Wang, Lu",
    editor = "Jurafsky, Dan  and
      Chai, Joyce  and
      Schluter, Natalie  and
      Tetreault, Joel",
    booktitle = "Proceedings of the 58th Annual Meeting of the Association for Computational Linguistics",
    month = jul,
    year = "2020",
    address = "Online",
    publisher = "Association for Computational Linguistics",
    url = "https://aclanthology.org/2020.acl-main.478/",
    doi = "10.18653/v1/2020.acl-main.478",
    pages = "5374--5386"
}

@inproceedings{liu-etal-2018-automatic,
    title = "Automatic Event Salience Identification",
    author = "Liu, Zhengzhong  and
      Xiong, Chenyan  and
      Mitamura, Teruko  and
      Hovy, Eduard",
    editor = "Riloff, Ellen  and
      Chiang, David  and
      Hockenmaier, Julia  and
      Tsujii, Jun{'}ichi",
    booktitle = "Proceedings of the 2018 Conference on Empirical Methods in Natural Language Processing",
    month = oct # "-" # nov,
    year = "2018",
    address = "Brussels, Belgium",
    publisher = "Association for Computational Linguistics",
    url = "https://aclanthology.org/D18-1154/",
    doi = "10.18653/v1/D18-1154",
    pages = "1226--1236"
}

@article{Sandhaus2008,
  author  = {Sandhaus, Evan},
  title   = {The New York Times Annotated Corpus},
  journal = {Linguistic Data Consortium, Philadelphia},
  year    = {2008},
  doi     = {10.35111/77ba-9x74}
}

@inproceedings{nenkova-passonneau-2004-evaluating,
    title = "Evaluating Content Selection in Summarization: The Pyramid Method",
    author = "Nenkova, Ani  and
      Passonneau, Rebecca",
    booktitle = "Proceedings of the Human Language Technology Conference of the North {A}merican Chapter of the Association for Computational Linguistics: {HLT}-{NAACL} 2004",
    month = may # " 2 - " # may # " 7",
    year = "2004",
    address = "Boston, Massachusetts, USA",
    publisher = "Association for Computational Linguistics",
    url = "https://aclanthology.org/N04-1019/",
    pages = "145--152"
}

\appendix
\section{System Prompts for LLMs} \label{sec:system prompts}
\begin{promptbox}[Baseline Prompt]
You are a linguist trained in identifying the most important pieces of a document. Given a document that has been split into sub-sentence units, which are numbered, you should identify the units that best describe the content of the document. 

\vspace{1cm}
Your response should take the form of a JSON object. You will provide the number for the chosen unit as a string.

\vspace{1cm}
Now let's try with a document:
\end{promptbox}

\begin{promptbox}[Summary-based Prompt]
You are an expert NLP system specialized in aligning document discourse units to summary sentences. Your task is to analyze a document and a summary, and for each document unit, determine whether it contributes to any information present in the summary.

\vspace{1cm}
For each unit:
1. Identify the specific summary sentence to which the unit may contribute information.
2. Compare the unit against that sentence: does it state, entail, paraphrase, vaguely relate to, or fail to support any part of it?
3. Include the unit only if it states, entails, or paraphrases part of the summary sentence.
4. Record the aligned part of the summary sentence in the triggers.
5. Decide whether the included unit is component, approximate, or duplicate.

\vspace{1cm}
Case: Normal Match vs. Non-Match

\vspace{1cm}
Rule:
For Normal Match, set approximate and component to false when the unit expresses the same core proposition as a portion of the summary. The unit does not need to exactly match the summary’s grammatical form. Differences in active/passive voice, tense, word order, or surface phrasing do not disqualify the unit, as long as the same information is clearly stated.
For Non-Match, exclude units that provide irrelevant context about an entity in the summary but do not express summary content directly.

\vspace{1cm}
Case: Fragments

\vspace{1cm}
Rule: 
Headlines, titles, fragments, or independent noun phrases are valid triggers if they assert summary content.

\vspace{1cm}
Case: Component

\vspace{1cm}
Rule: 
Set component to true when the unit provides one item, detail, or subclaim from a larger set that the summary combines into a broader statement, such as when several document units are summarized together as a list or category.

\vspace{1cm}
Case: Approximate

\vspace{1cm}
Rule:
Set approximate to true when the unit is a valid match but expresses the summary content indirectly, incompletely, or with a meaningful shift in specificity, implication, or framing.

\vspace{1cm}
Case: Duplicate 

\vspace{1cm}
Rule:
Fill the duplicate list with the ID of the earliest unit that triggers the same summary content when the current unit expresses the same information redundantly.
\end{promptbox}

\section{Expected Output for Summary-based Prompts} \label{output}
    \{
        "id": "74",
        "approximate": false,
        "component": false,
        "duplicate": ["70"],
        "triggers": "period of limited exposure generally leads to attrition"
    \}

\section{RST example}\label{sec:rst}

\begin{figure}[h!t]
    \centering
    \includegraphics[width=\columnwidth, trim={1.6cm 16.5cm 7cm 4cm}, clip=true]{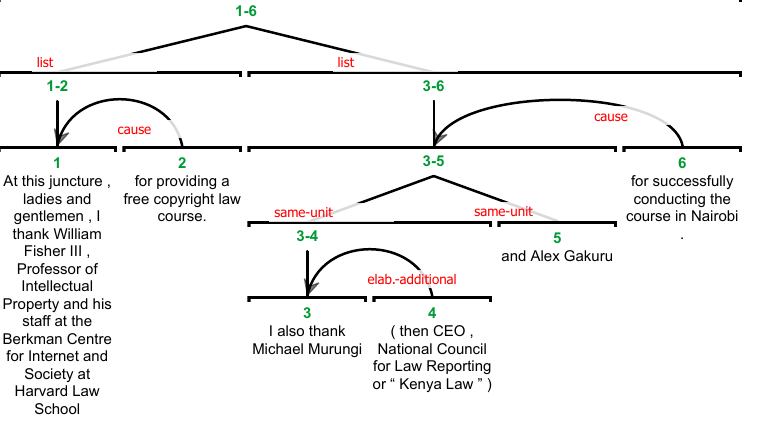}
    \caption{Fragment from an RST tree in the GUM corpus.}
    \label{fig:rst}
\end{figure}

RST trees partition documents into Elementary Discourse Units (EDUs), which generally correspond to clauses, but may be discontinuous. Less central units form satellites for more central units (their `nuclei'), and equally prominent units may be coordinated as multinuclear nodes. The discourse function of each unit is expressed using a category label, and constituents containing multiple unit can recursively serve as satellites or nuclei in larger subtrees, recursively. Figure \ref{fig:rst} illustrates this in an RST tree fragment.

The most central units are the coordinated constituents [1-2] and [3-6], which form a list of `thanking' propositions. Unit [2] is a satellite of unit [1], expressing \textsc{cause} for the thanks, and similarly unit [6] is a \textsc{cause} satellite of [3-5]. Units [3] and [5] correspond to a discontinuous proposition, indicated by the \textsc{same-unit} label. The proposition is interrupted by unit [4], an elaboration satellite giving more information about a person being thanked.

\section{One-shot Demonstration Pairings} 
See Table~\ref{tab:test-in-genre-examples} for one-shot demonstration document-summary pair from the dev set and the number of subtypes included in each demonstration.

\label{oneshot_demo}
\begin{table*}[t]
\centering
\resizebox{\textwidth}{!}{%
\begin{tabular}{llrrrrr}
\toprule
Genre & Dev example (summary) & Match & Approx. & Duplicate & Component & Types \\
\midrule
academic     & librarians (1)     &  7 & 4 & 1 &  0 & 3 \\
bio          & byron (2)          &  4 & 6 & 2 &  3 & 4 \\
conversation & grounded (3)       &  4 & 7 & 4 &  2 & 4 \\
court        & loan (5)           &  6 & 4 & 4 &  3 & 4 \\
essay        & tools (4)          &  2 & 5 & 2 &  2 & 4 \\
fiction      & beast (2)          &  4 & 7 & 2 &  3 & 4 \\
interview    & gaming (1)         &  8 & 1 & 2 &  8 & 4 \\
letter       & wiki (5)           & 10 & 2 & 3 &  1 & 4 \\
news         & homeopathy (1)     &  5 & 4 & 4 &  3 & 4 \\
podcast      & bangladesh (2)     &  7 & 3 & 3 &  4 & 4 \\
reddit       & macroeconomics (2) &  1 & 4 & 1 & 10 & 4 \\
speech       & inauguration (5)   &  6 & 4 & 4 &  8 & 4 \\
textbook     & labor (2)          &  4 & 8 & 6 &  0 & 3 \\
vlog         & portland (1)       &  8 & 6 & 5 &  2 & 4 \\
voyage       & coron (3)          &  5 & 2 & 2 &  0 & 3 \\
whow         & joke (2)           &  1 & 1 & 1 &  6 & 4 \\
\bottomrule
\end{tabular}%
}
\caption{Development-set examples used for one-shot in-genre prompting
on the test set. Document names omit the prefix
\texttt{GUM\_<genre>\_}. Types counts
categories with a nonzero value.}
\label{tab:test-in-genre-examples}
\end{table*}

\section{Number of Original versus Normalized Qwen outputs} 
See Table~\ref{tab:qwen_test_normalization} for the number of documents in Qwen outputs that require normalization.

\label{qwennormalization}
\begin{table}[t]
\centering
\resizebox{\columnwidth}{!}{%
\begin{tabular}{lrrr}
\toprule
Prompting & Normalized & Unnormalized & Total \\
\midrule
Zero-shot & 58 & 102 & 160 \\
One-shot  & 11 & 149 & 160 \\
\midrule
Total     & 69 & 251 & 320 \\
\bottomrule
\end{tabular}%
}
\caption{Qwen test-set prediction files by whether at least one predicted
EDU ID required normalization.}
\label{tab:qwen_test_normalization}
\end{table}

\section{Fine-grained evaluation} \label{finegrainedeval}

Table \ref{tab:fine-grained-results} provides numbers for the fine grained evaluation using exact subtypes (including prediction of \textcolor{cyan}{approximate} and \textcolor{OliveGreen}{component} status). In this exact evaluation setting, scores are much lower.

\begin{table}[t]
    \centering
    \setlength{\tabcolsep}{5pt}
    \renewcommand{\arraystretch}{1.2}
    \begin{tabular}{@{}llccc@{}}
        \toprule
        Setting & Model & P & R & F1 \\
        \midrule
        \multirow{2}{*}{\shortstack[l]{Summary\\zero-shot}}
            & GPT & 0.323 & 0.509 & 0.395 \\
        \cmidrule(l){2-5}
            & Claude & 0.345 & 0.256 & 0.294 \\
        \midrule
        \multirow{2}{*}{\shortstack[l]{Summary\\one-shot}}
            & GPT & 0.395 & 0.479 & 0.433 \\
        \cmidrule(l){2-5}
            & Claude & 0.411 & 0.328 & 0.365 \\
        \bottomrule
    \end{tabular}
    \caption{Precision (P), recall (R), and F1 for the subtype-aware
    fine-grained evaluation scheme. Extra predicted EDUs are false
    positives, while a salient EDU assigned the wrong subtype is counted
    once as a false negative.}
    \label{tab:fine-grained-results}
\end{table}

\section{Salience Score} 
\begin{figure*}[t]
    \centering
    \includegraphics[width=\textwidth]{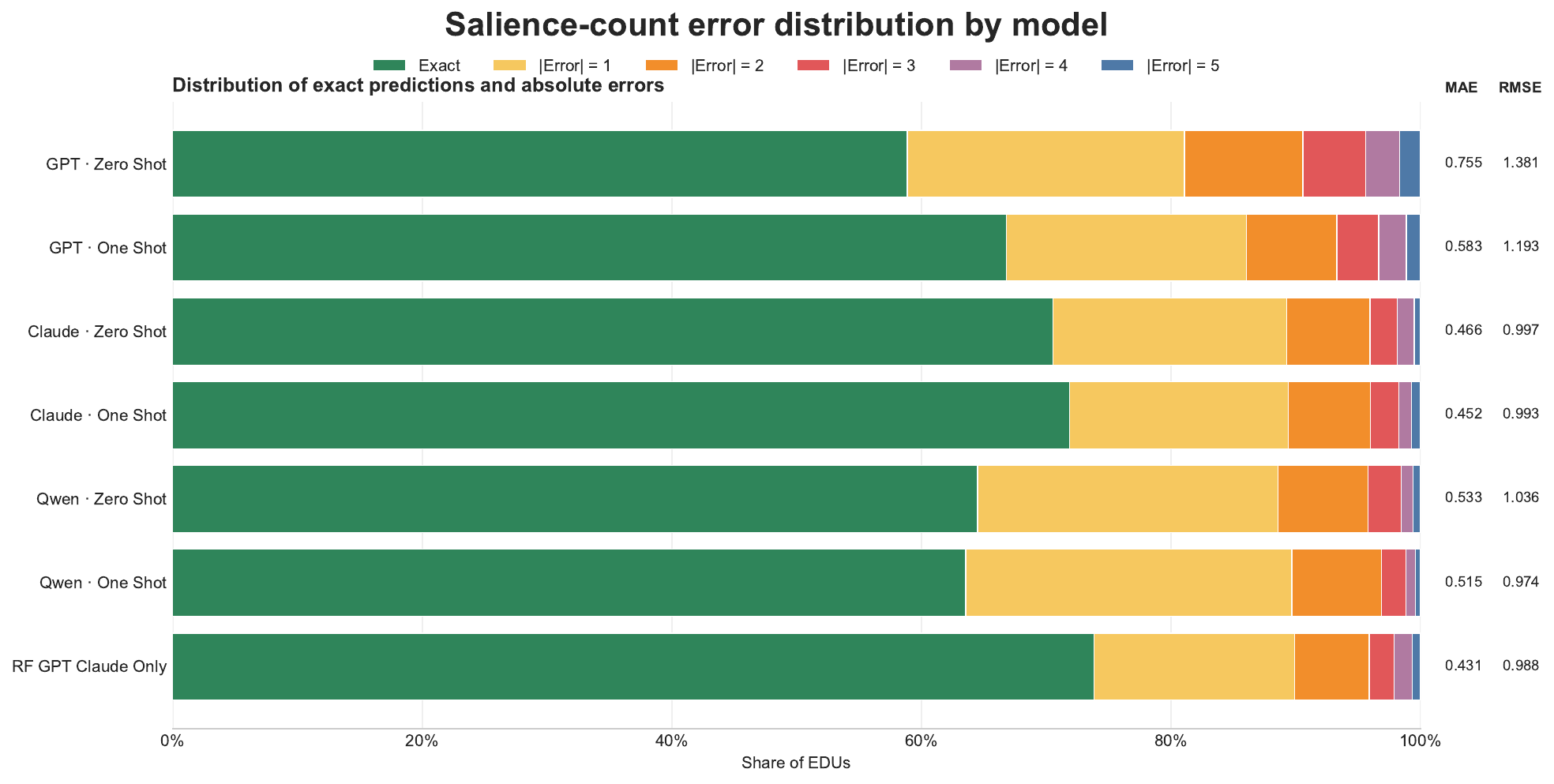}
    \caption{Absolute error distance from gold salience by model. Qwen models have more errors than any model except GPT zero shot, but disproportionately many off-by-one errors, usually between 0 and 1, resulting in better RMSE than expected, but worse MAE. Claude one-shot is the best model by F1 score except for the ensemble.}
    \label{fig:saliencescore}
\end{figure*}

Figure~\ref{fig:saliencescore} shows the frequency distribution of prediction errors with absolute magnitudes ranging from 0 to 5 across different models.

\label{saliencescore}

\section{Ensemble details}
\label{sec:app-ensemble}

\subsection{Finetuning}
\label{sec:ens-tuning}
All tuning is done using the \texttt{train} partition with 5-fold cross validation. Folds are assigned at the document level to prevent leakage of EDUs. Non-salient EDUs are significantly more common than salient; each parameter configuration is scored by the F1 of the positive class on the out-of-fold predictions rather than per fold. Categorical features are one-hot encoded. We use Optuna's Tree-structured Parzen Estimator (TPE) with 60 trails per model.
\subsubsection{Random Forest}
The search space for the Random Forest model is outlined in Table \ref{tab:rf_search_space}.

\begin{table}[t]
    \centering
    \small
    \setlength{\tabcolsep}{5pt}
    \renewcommand{\arraystretch}{1.15}
    \begin{tabularx}{\columnwidth}{@{}lX@{}}
        \toprule
        \textbf{Parameter} & \textbf{Search space} \\
        \midrule
        Number of trees & $\{200, 300, \ldots, 800\}$ \\
        Max feats per split & $\{\text{sqrt}, \text{log2}, 0.3, 0.5, 0.7\}$ \\
        Min samples per leaf & Integers in $[1,20]$ \\
        Imbalance strategy & \{none, class weight, balanced subsample\} \\
        Positive-class weight & Log-uniform on $[1, \max(1.5r, 2)]$ \\
        Threshold & Uniform on $[0.1, 0.6]$ \\
        \bottomrule
    \end{tabularx}
    \caption{Hyperparameter search space for the Random Forest model.}
    \label{tab:rf_search_space}
\end{table}
The final parameter configuration for this model is:
400 trees, max features 0.3, min samples per leaf 1, class weighting with a positive-class weight of 1.90, and a threshold of 0.465, using 27 features. The final model is retrained on the full \texttt{train} set and applied to the test partition with this threshold.

\subsubsection{XGBoost}

The search space for the XGBoost model is outlined in Table \ref{tab:xgb_search_space}.

\begin{table}[t]
    \centering
    \begin{tabular}{@{}ll@{}}
        \toprule
        \textbf{Parameter} & \textbf{Search space} \\
        \midrule
        Max depth & Integers in $[3, 10]$ \\
        Learning rate & Log-uniform on $[0.01, 0.3]$ \\
        Row subsample & Uniform on $[0.5, 1.0]$ \\
        Column subsample & Uniform on $[0.5, 1.0]$ \\
        Min child weight & Integers in $[1, 10]$ \\
        Boosting rounds & $\{100, 150, \ldots, 600\}$ \\
        \texttt{scale\_pos\_weight} & Uniform on $[0.5r, 1.5r]$ \\
        Threshold & Uniform on $[0.1, 0.6]$ \\
        \bottomrule
    \end{tabular}
    \caption{Hyperparameter search space for the XGBoost model.}
    \label{tab:xgb_search_space}
\end{table}

The final parameter configuration for this model is:
Max depth 9, learning rate 0.054, row subsample 0.864, column subsample 0.750, min child weight 9, 300 boosting rounds, scale\_pos\_weight 8.51, and a threshold of 0.546, using 34 features. The final model is retrained on the full train set and applied to the test partition with this threshold.

\subsection{Features}
Table \ref{tab:all-features} contains all of the features considered during ensemble training. Fine-tuning the ensemble models included pruning out less helpful features via greedy backward elimination, specifically using our Random Forest model. We provided the models with all possible features to begin. Features were ranked by decrease in impurity, and the removal of each was tested to investigate whether the F1 score improved. The features which remained are bolded in the table.

\begin{figure*}[t]
    \centering
    \includegraphics[width=\textwidth]{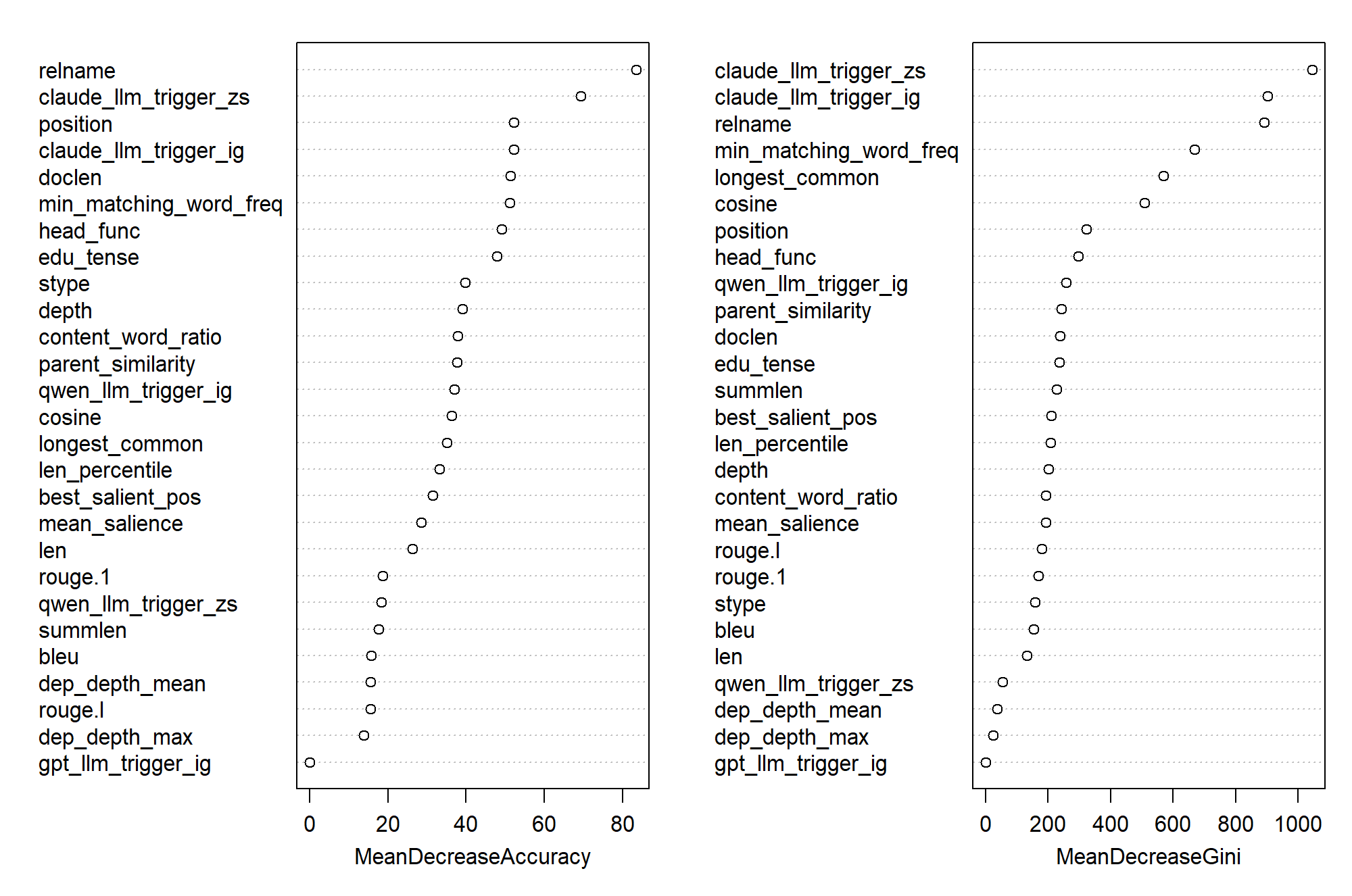}
    \caption{Variable importance for Random Forest, calculated on \texttt{dev}.}
    \label{fig:rf_varimp}
\end{figure*}

\begin{figure}[htbp]
    \centering
    \includegraphics[width=0.5\textwidth]{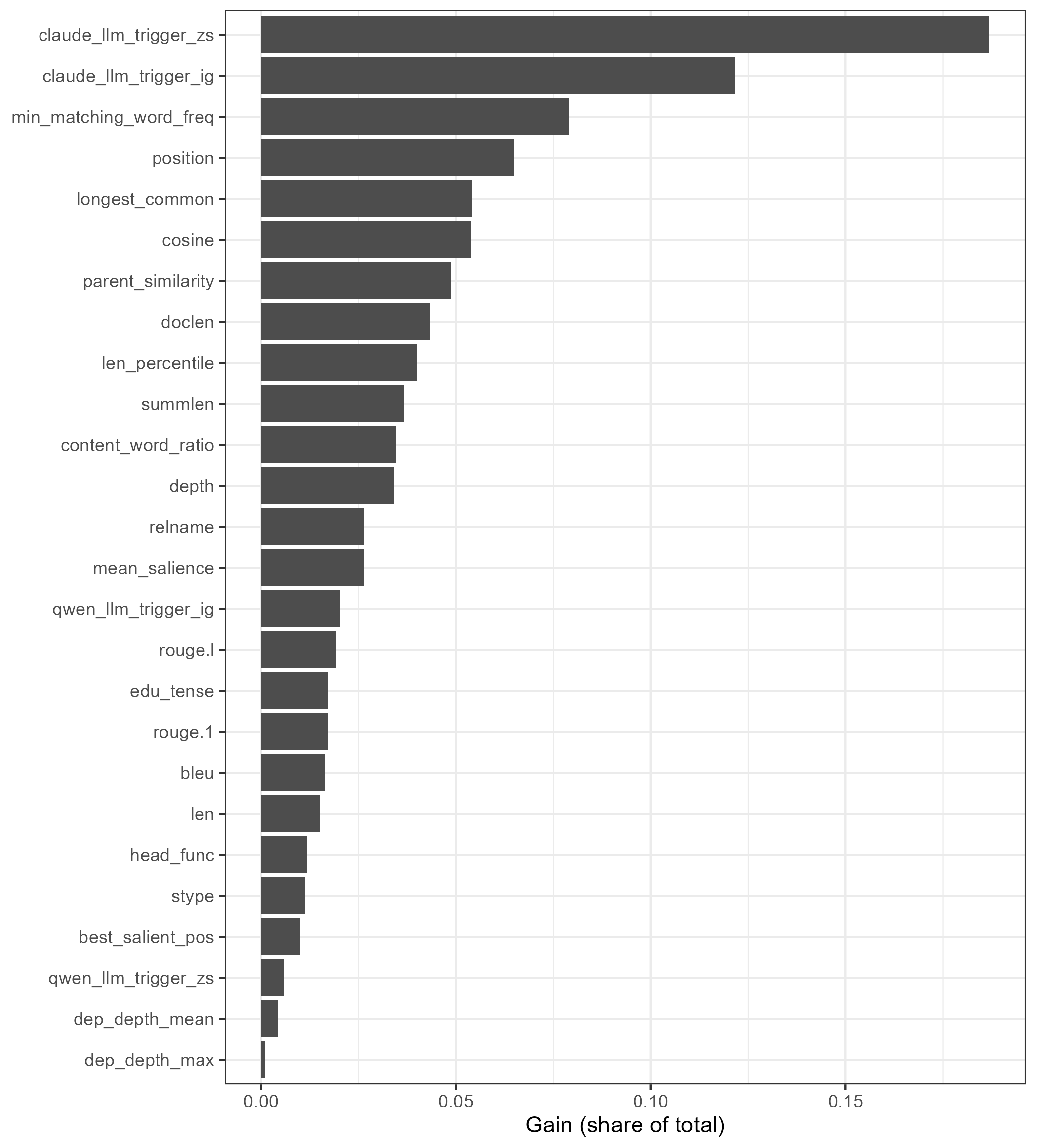}
    \caption{Variable importance for XGBoost, calculated on \texttt{dev}.}
    \label{fig:xgb_varimp}
\end{figure}

Using these final features, feature importance for the Random Forest model and XGBoost models are  reported in Figures \ref{fig:rf_varimp} and \ref{fig:xgb_varimp}, respectively.

\label{sec:ens-features}
\begin{table*}[t]
\centering
\small
\begin{tabular}{@{}l p{10cm}@{}}
\toprule
\textbf{Feature} & \textbf{Description} \\
\midrule
\textbf{\texttt{position}} & Normalized position of the EDU within its document \\
\textbf{\texttt{depth}} & Normalized depth of the EDU in the discourse (RST) tree \\
\textbf{\texttt{parent\_similarity}} & Similarity between the EDU and its parent EDU \\
\textbf{\texttt{len}} & Token length of the EDU \\
\textbf{\texttt{relname}} & RST discourse relation connecting the EDU to its parent \\
\texttt{mention\_count} & Number of entity mentions whose span overlaps EDU \\
\texttt{unique\_salient} & Number of overlapping salient mentions that belong to a singleton coreference cluster\\
\textbf{\texttt{mean\_salience}} & Mean salience score of the entities tagged as salient \\
\textbf{\texttt{bleu}} & BLEU overlap between the EDU and the summary text \\
\textbf{\texttt{rouge-1}} & ROUGE-1 F1 between the EDU and the summary text \\
\texttt{rouge-2} & ROUGE-2 F1 between the EDU and the summary text \\
\textbf{\texttt{rouge-l}} & ROUGE-L F1 between the EDU and the summary text \\
\textbf{\texttt{cosine}} & Cosine similarity between EDU and summary embeddings \\
\textbf{\texttt{head\_func}} & Dependency function of the EDU's syntactic head (e.g.\ \textit{nsubj}, \textit{root}) \\
\textbf{\texttt{edu\_tense}} & Grammatical tense of the EDU \\
\textbf{\texttt{stype}} & Sentence type of the EDU's containing sentence \\
\textbf{\texttt{best\_salient\_pos}} & POS tag of the head token of the highest-ranked salient mention overlapping the EDU \\
\texttt{is\_sent} & Whether the EDU corresponds to a full sentence (binary) \\
\textbf{\texttt{doclen}} & Token length of the source document\\
\texttt{summlen} & Length of the reference summary \\
\textbf{\texttt{longest\_common}} & Length of the longest common substring between the EDU and the summary \\
\textbf{\texttt{min\_matching\_word\_freq}} & Minimum corpus frequency among words shared between the EDU and the summary \\
\texttt{cosine\_norm} & Normalized cosine similarity between EDU and summary \\
\textbf{\texttt{len\_percentile}} & Percentile rank of the EDU's length relative to other EDUs in the document \\
\textbf{\texttt{content\_word\_ratio}} & Proportion of content tokens in the EDU \\
\textbf{\texttt{dep\_depth\_mean}} & Mean depth of the EDU's dependency parse \\
\textbf{\texttt{dep\_depth\_max}} & Maximum depth of the EDU's dependency parse \\
\texttt{entity\_cohort\_size} & Size of the largest coreference chain among entities mentioned in the EDU\\
\textbf{\texttt{claude\_llm\_trigger\_zs}} & Binary indicator that an LLM (Claude) tagged the EDU as salient, zero-shot \\
\texttt{gpt\_llm\_trigger\_zs} & Binary indicator that an LLM (GPT) tagged the EDU as salient, zero-shot\\
\textbf{\texttt{qwen\_llm\_trigger\_zs}} & Binary indicator that an LLM (Qwen) tagged the EDU as salient, zero-shot \\
\textbf{\texttt{claude\_llm\_trigger\_ig}} & Binary indicator that an LLM (Claude) tagged the EDU as salient, one-shot \\
\textbf{\texttt{gpt\_llm\_trigger\_ig}} & Binary indicator that an LLM (GPT) tagged the EDU as salient, one-shot\\
\textbf{\texttt{qwen\_llm\_trigger\_ig}} & Binary indicator that an LLM (Qwen) tagged the EDU as salient, one-shot \\
\bottomrule
\end{tabular}
\caption{All possible features used for EDU-level salience classification. Features kept after pruning on \texttt{dev} are bolded.}
\label{tab:all-features}
\end{table*}

\section{Position of Errors Across Documents}
\label{sec:error-position}
\begin{figure*}[htpb]
  \centering
  \includegraphics[width=1.0\textwidth]{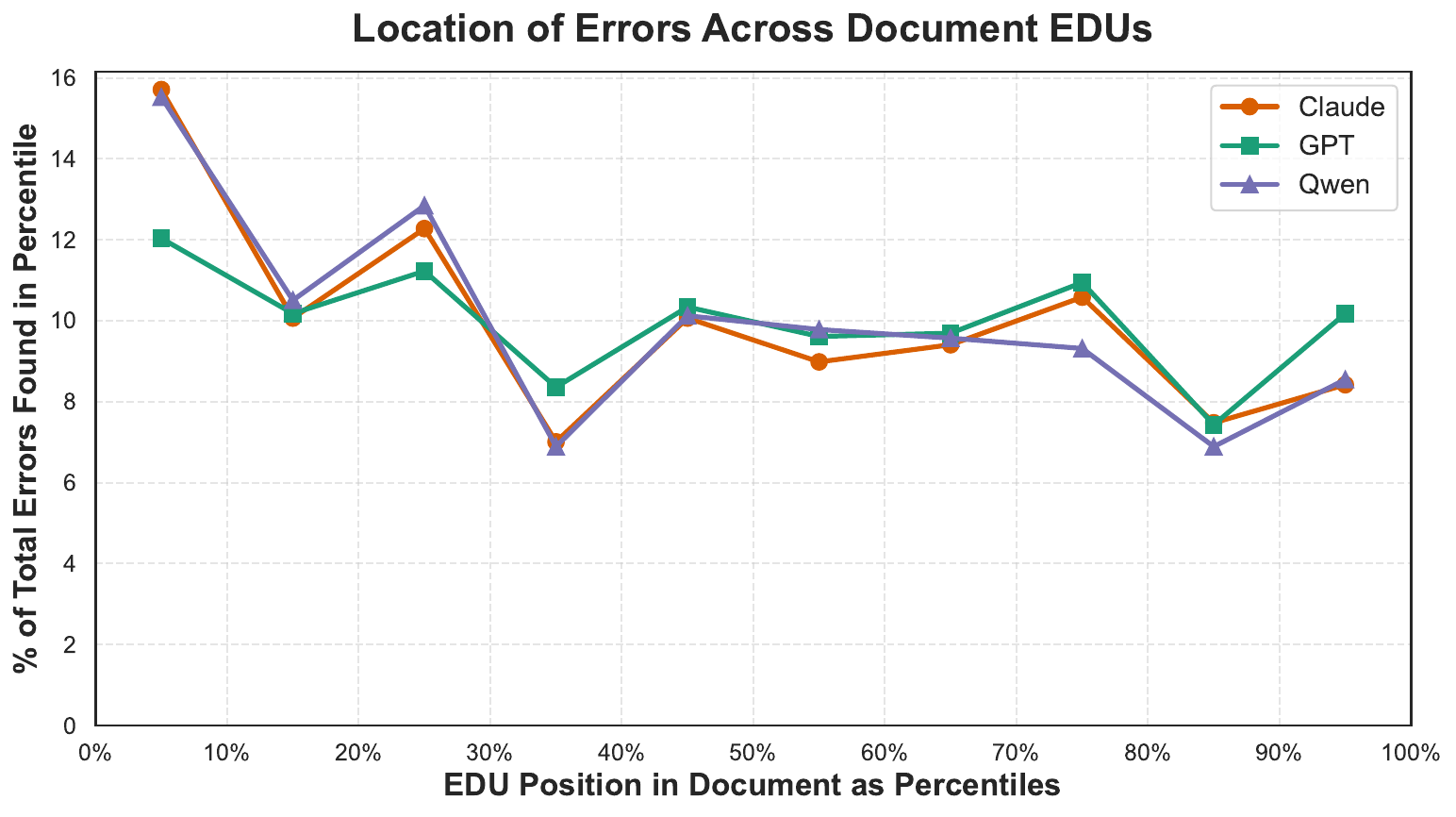}
  \caption{Distribution of percentage of errors across EDUs in a document.}
  \label{fig:error-position}
\end{figure*}

When looking at the distribution of errors across the length of documents in Figure \ref{fig:error-position}, we expected to find a lost-in-the-middle effect. Models often focus more on the beginning or end of documents, but we did not observe this pattern. Across all models, the errors were concentrated in the first third of a document. This effect was especially notable in Claude and Qwen, with GPT making slightly fewer errors at the beginning. GPT showed a more stable distribution of errors, fluctuating between approximately 8\% and 12\%. 

\begin{figure*}[htpb]
\centering
  \includegraphics[width=1.0\textwidth]{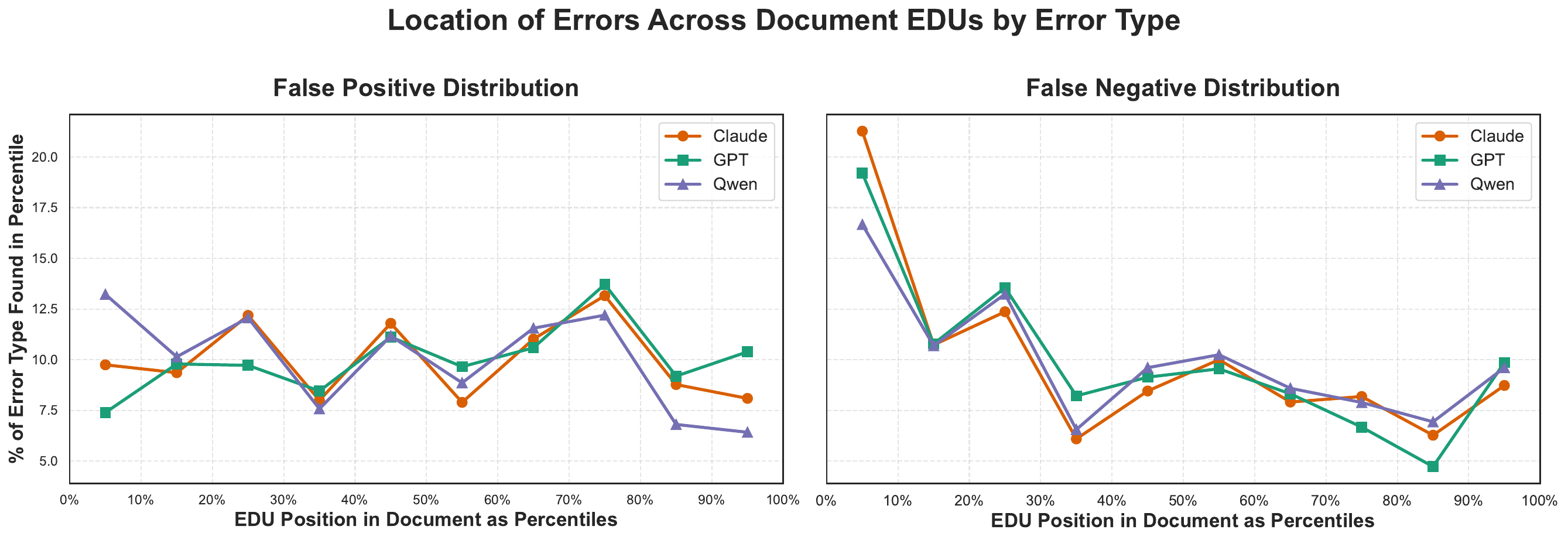}
  \caption{Distribution of percentage of errors across EDUs in a document by error type.}
  \label{fig:error_position-type}
\end{figure*}

Splitting the graphs between false positives and false negatives, as shown in Figure \ref{fig:error_position-type}, revealed that most of the errors occurring at the beginnings of documents were false negatives. This indicates that the models were not sufficiently attending to information at the beginnings of documents that humans considered salient. In contrast, the number of false negatives was, on average, lower but fluctuated more from percentile to percentile. The most notable spike was right before the ends of documents. Overall, models did not show a lost-in-the-middle effect and underpredicted the number of salient EDUs at the beginnings of documents and overpredicted EDUs later in documents,

\end{document}